\documentclass[conference]{IEEEtran}
\IEEEoverridecommandlockouts

\usepackage[T1]{fontenc}
\usepackage{subcaption}
\usepackage{graphicx}
\usepackage{amsmath, amssymb}
\usepackage[nolist]{acronym}

\begin{acronym}[XXX] % Give the longest label here so that the list is nicely aligned
\acro{ai}[AI]{Artificial Intelligence}
\acro{auc}[AUC]{Area Under the Curve}
\acro{bva}[BVA]{Best Validation Accuracy}
\acro{cnn}[CNN]{Convolutional Neural Network}
\acro{dl}[DL]{Deep Learning}
\acro{dnn}[DNN]{Deep Neural Network}
\acro{fl}[FL]{Federated Learning}
\acro{ff}[FF]{Feedforward Neural Network}
\acro{gelu}[GELU]{Gaussian Error Linear Unit}
\acro{glai}[GLAI]{GreenLightningAI}
\acro{iid}[\textit{i.i.d.}]{independent and identically distributed}
\acro{kd}[KD]{Knowledge Distillation}
\acro{ml}[ML]{Machine Learning}
\acro{mlp}[MLP]{Multilayer Perceptron}
\acro{mva}[MVA]{Maximum Validation Accuracy}
\acro{mha}[MHA]{Multihead Attention}
\acro{moe}[MoE]{Mixture-of-Experts}
\acro{mimick}[NM]{Neural Mimicking}
\acro{nlp}[NLP]{Natural Language Processing}
\acro{cv}[CV]{Computer Vision}
\acro{nn}[NN]{Neural Network}
\acro{oui}[OUI]{Overfitting-Underfitting Indicator}
\acro{prelu}[PReLU]{Parametric Rectified Linear Unit}
\acro{poc}[PoC]{Proof-of-Concept}
\acro{qk}[QK]{Quantitative Knowledge}
\acro{relu}[ReLU]{Rectified Linear Unit}
\acro{silu}[SiLU]{Sigmoid Linear Unit}
\acro{sgd}[SGD]{Stochastic Gradient Descent}
\acro{svd}[SVD]{Singular Value Decomposition}
\acro{sota}[SOTA]{state-of-the-art}
\acro{sk}[SK]{Structural Knowledge}
\acro{tl}[TL]{Transfer Learning}
\acro{vt}[VT]{Vision Transformer}
\acro{vit}[ViT]{Vision Transformer}
\acro{wd}[WD]{Weight Decay}
\end{acronym}

\begin{document}
\title{Detecting Atypical Clients in Federated Learning via Representation-Level Divergence}
%\titlerunning{Detecting Atypical Clients}

\author{
\makebox[.33\linewidth][c]{
\begin{tabular}{c}
Cristian Pérez-Corral*\thanks{*Corresponding author} \\
\textit{Universitat Politècnica de València} \\
Valencia, Spain \\
cpercor@upv.es
\end{tabular}
}
\makebox[.33\linewidth][c]{
\begin{tabular}{c}
Jose I. Mestre \\
\textit{Universitat Politècnica de València} \\
Valencia, Spain \\
jimesmir@disca.upv.es
\end{tabular}
}

\makebox[.33\linewidth][c]{
\begin{tabular}{c}
Alberto Fernández-Hernández \\
\textit{Universitat Politècnica de València} \\
Valencia, Spain \\
a.fernandez@upv.es
\end{tabular}
}\\[1em]
\makebox[.33\linewidth][c]{
\begin{tabular}{c}
Manuel F. Dolz \\
\textit{Universitat Jaume I} \\
Castelló de la Plana, Spain \\
dolzm@uji.es
\end{tabular}
}
\makebox[.33\linewidth][c]{
\begin{tabular}{c}
Enrique S. Quintana-Ortí \\
\textit{Universitat Politècnica de València} \\
Valencia, Spain \\
quintana@disca.upv.es
\end{tabular}
}
}
\maketitle              % typeset the header of the contribution
\begin{abstract}
Federated learning enables collaborative training across distributed clients with heterogeneous data, but such heterogeneity often leads to unstable updates and degraded global performance. Moreover, in practical deployments, client updates may deviate from the expected behavior not only due to benign not \acl{iid} distributions, but also due to distributional shifts or anomalous inputs, raising concerns about the reliability of the aggregation process.
In this work, we propose a lightweight geometric signal to quantify the functional deviation of a client with respect to the global model. Instead of comparing model parameters or gradients, our approach measures how the local training of each client alters the activation-induced partition of the input space, evaluated on a shared probe set. This yields a permutation-invariant, interpretable metric of client--global divergence that captures differences in how data is processed by the model.
We show that this signal effectively identifies clients that induce atypical functional changes, distinguishing stable yet heterogeneous clients from those whose updates significantly diverge from the global regime. As a result, the proposed metric provides a simple tool for monitoring client behavior and enabling risk-aware aggregation strategies in federated learning systems.

\end{abstract}
\section{Introduction}
The proliferation of connected devices and the ongoing digitalization of organizations have led to an unprecedented growth in the amount of data that must be processed efficiently and securely~\cite{chen2014bigdata}. Centralizing such data is often inefficient, vulnerable, and impractical due to privacy, legal, and operational constraints. In this context, \ac{fl} has emerged as a paradigm for collaboratively training \ac{dl} models while keeping data local to each participant, providing inherent advantages in terms of privacy and security~\cite{kairouz2019,mcmahan2017}.

However, \ac{fl} introduces fundamental challenges compared to centralized learning. In particular, data across clients is typically heterogeneous, which leads to discrepancies in local training dynamics and complicates model aggregation. This heterogeneity can manifest as feature shift or label skew, where different clients exhibit distinct input distributions or class proportions, making it difficult to learn a globally consistent model~\cite{mcmahan2017}.

Beyond statistical heterogeneity, the distributed nature of \ac{fl} also introduces additional concerns from a security and reliability perspective. In realistic deployments, clients operate in dynamic and potentially untrusted environments, where data distributions may evolve over time or deviate from expected conditions due to sensor drift, environmental changes, or even corrupted or anomalous inputs. As a result, the updates contributed by each client may not only reflect benign heterogeneity, but may also exhibit atypical behavior that can negatively impact the global model.

From a system perspective, this raises an important challenge: distinguishing between expected variability due to non-\ac{iid} data and potentially harmful deviations that may compromise the stability or trustworthiness of the training process. Traditional approaches often rely on comparing model parameters or gradients, which may fail to capture how such deviations affect the functional behavior of the model. In particular, they provide limited insight into how different clients induce changes in the internal representation and decision boundaries learned by the \ac{nn}.

Motivated by these limitations, we adopt a complementary perspective centered on the geometric structure induced by \acp{nn}. In particular, we focus on how different clients organize the input space through their activation patterns, and how these differences can be leveraged to better understand heterogeneity and detect atypical behaviors in federated settings. From this standpoint, the main contributions of this work can be summarized as follows:
\begin{itemize}
    \item We introduce a geometric perspective to analyze client behavior in \ac{fl}, based on the partition of the input space induced by activation patterns in \acp{nn}.
    
    \item We propose a lightweight metric to quantify the functional divergence between client and global models, capturing differences in how inputs are organized and processed beyond parameter-level comparisons.
    
    \item We show that this metric provides an interpretable signal to monitor data heterogeneity and to identify atypical client contributions in federated learning systems.
\end{itemize}

The remainder of the paper is organized as follows. Section~\ref{sec:related} reviews related work on \ac{nn} structure and \ac{fl} security and privacy. Section~\ref{sec:geometric} introduces the necessary definitions and the proposed metric. Section~\ref{sec:experiments} evaluates the metric on two different models and demonstrates its usefulness. Section~\ref{sec:conclusion} concludes the paper and discusses future research directions.

\section{Related Work}
\label{sec:related}
\ac{fl} enables collaborative learning across distributed data holders without transferring raw data, typically through iterative rounds of local optimization and server-side aggregation~\cite{mcmahan2017}. In practice, a major obstacle is the statistical heterogeneity of client data, which often leads to unstable convergence and degraded global performance. Classical approaches such as \textsc{FedProx} mitigate excessive local deviation by introducing a proximal term~\cite{li2020}, while adaptive server-side optimization methods such as \textsc{FedOpt} improve robustness under heterogeneous data distributions~\cite{reddi2021}. Other works address specific forms of non-\ac{iid}ness, such as feature shift affecting batch-normalization statistics (\textsc{FedBN})~\cite{xiaoxiao2021}, or personalization-oriented methods that decompose the model into shared and client-specific components, such as \textsc{FedRep} and \textsc{FedBABU}~\cite{collins2021,oh2022}. In contrast to these approaches, our goal is not to modify the optimization procedure itself, but to provide a lightweight signal that characterizes how much a client functionally deviates from the global model.

A complementary line of work studies \acs{nn} through the geometry induced by their activation patterns. In \ac{relu}-based \acs{nn}, sign patterns define a partition of the input space into linear or piecewise-affine regions, whose number and geometry are closely related to expressivity and representation structure~\cite{montufar2014number,raghu2017expressive}. Recent works have begun to analyze the dynamics of such patterns during training. Xu et al.~\cite{xu2024127174} study convergence properties once activations are fixed, whereas Hartmann et al.~\cite{hartmann} analyze activation statistics such as entropy and similarity. 

    From a security and privacy viewpoint, \ac{fl} is often motivated by keeping raw data local, but this alone does not guarantee either privacy or integrity. On the privacy side, secure aggregation protocols aim to prevent the server from inspecting individual client updates by only revealing their aggregate sum~\cite{bonawitz2017}. On the integrity side, Byzantine-robust aggregation methods such as \textsc{Krum} seek to tolerate arbitrary or adversarial client updates during distributed optimization~\cite{blanchard2017}. More broadly, this line of work highlights that distributed learning systems require mechanisms not only for optimization, but also for assessing whether client contributions are trustworthy or atypical. Our proposal is complementary to these approaches: rather than providing cryptographic privacy or formal Byzantine robustness, we introduce a geometric monitoring signal that quantifies how strongly a client update deviates from the functional regime induced by the global model. This makes it particularly relevant for settings where heterogeneity, drift, or anomalous behavior may compromise the reliability of aggregation.

\section{Geometric Client Divergence}
\label{sec:geometric}
\begin{figure*}[!t]
    \centering
    \includegraphics[width=1\linewidth]{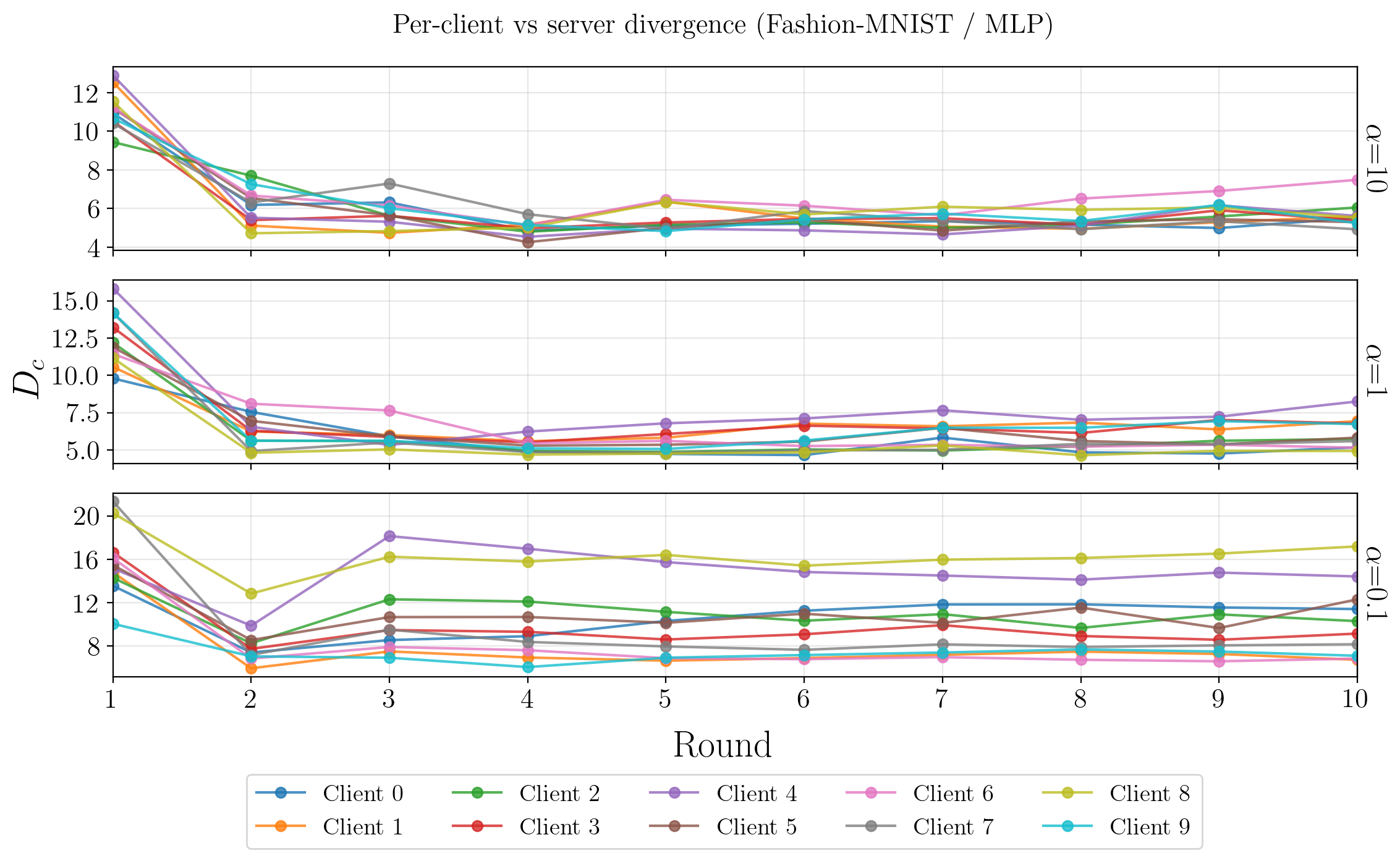}
    \caption{Per-client $D_c$ across rounds for different Dirichlet $\alpha$'s. Higher heterogeneity ($\alpha=0.1$) results in increased divergence and persistent client deviations, while near \ac{iid} settings exhibit stable convergence.}
    \label{fig:exp1}
\end{figure*}
In this section, we propose a metric to quantify the geometric difference between two models in a federated learning setting. Recent work has shown that, in general, activation patterns (and therefore the induced geometric partition of the input space) tend to stabilize earlier than the model parameters during training~\cite{perezcorral2026}. Under this perspective, models trained on compatible data distributions are expected to organize the input space in similar ways. We leverage this intuition to assess whether a client deviates excessively from the global model in terms of its induced geometric structure.

Let $f_\theta$ be a \ac{relu} \ac{ff} with $L$ layers. For an input $x \in \mathbb{R}^d$, denote by
\(
z_l(x;\theta) \in \mathbb{R}^{n_l}
\) the pre-activation vector at layer $l$, and define the corresponding binary activation pattern
\[
a_l(x;\theta) = \mathbf{1}\{ z_l(x;\theta) > 0 \} \in \{0,1\}^{n_l},
\]
where the inequality is applied element-wise. Each layer induces a partition of the input space into regions characterized by activation patterns; those samples with the same activation pattern belong to the same region. These regions define a piecewise-affine decomposition of the function computed by $f_\theta$. Let $P = \{x_1, \dots, x_m\}$ be a fixed probe set. We study the geometry induced by the model on $P$ through the activation patterns. To do so, we propose a new metric: for a given layer $l$, we define the \textit{activation affinity} between two inputs $x_i, x_j \in P$ as
\[
K_l^\theta(i,j)
=
1 - \frac{1}{n_l} H\big(a_l(x_i;\theta), a_l(x_j;\theta)\big),
\]
where $H(\cdot,\cdot)$ denotes the Hamming distance. This metric quantifies the ``distance'' (in terms of number of hyperplanes to be crossed) between the samples $x_i$ and $x_j$, under the parameters $\theta$. The matrix $K_l^\theta \in [0,1]^{m \times m}$ captures how the layer groups inputs, where a high value in entry $(i,j)$ indicates that the corresponding inputs lie in the same or nearby regions, while low values indicate separation by multiple activation hyperplanes. 

The main idea is to compare how two different models organize the probe samples pairwise. The intuition is that similar models (or at least, not divergent ones) should respect the activation affinity between samples. Given two models $f_\theta$ and $f_{\theta'}$, we define their \textit{layer-wise geometric affinity} as
\[
d_l(\theta,\theta') = \| K_l^\theta - K_l^{\theta'} \|_F,
\]
where $\|\cdot\|_F$ denotes the Frobenius norm. In other words, two models are considered similar if their layers group samples at comparable distances in terms of activation affinity. \acsp{nn} induce a hierarchical partition of the input space, where early layers define coarse separations and deeper layers refine them. To account for this structure, we define a hierarchical divergence that prioritizes discrepancies at early layers, in the sense that differences at shallow layers indicate fundamentally different partitions of the input space, while discrepancies at deeper layers are only informative when earlier partitions are similar. Let \(\lambda_l = \exp(-\beta \, d_l(\theta,\theta'))\), with $\beta > 0$. We define the \textit{(hierarchical) geometric divergence} as
\[
D_{\mathrm{hier}}(\theta,\theta')
=
d_1(\theta,\theta')
+
\sum_{l=2}^L
\left(
\prod_{r=1}^{l-1} \lambda_r
\right)
d_l(\theta,\theta').
\]

This formulation ensures that discrepancies at deeper layers are progressively downweighted when earlier layers already exhibit significant differences.

In a \ac{fl} setting, let $\theta_g^{(t)}$ be the global model at round $t$, and $\theta_c^{(t)}$ the model obtained by client $c$ after local training. Therefore, the geometric divergence of the client with respect to the central server follows
\[
D_c^{(t)} = D_{\mathrm{hier}}(\theta_g^{(t)}, \theta_c^{(t)}).
\]

This quantity measures how much the local training of client $c$ alters the functional geometry of the global model, providing an interpretable signal of client deviation.
\section{Experiments}
\label{sec:experiments}
In this section we report the empirical results of our proposal, evaluating whether the proposed metric effectively captures geometrical divergence and can be used as a signal to identify potentially problematic clients during aggregation.
 To this end, we consider two federated scenarios:
\begin{itemize}
    \item An \ac{mlp} on Fashion-MNIST with a hidden dimension of 128 units.
    \item A ResNet-18 on CIFAR-10.
\end{itemize}
Both models are trained using \ac{sgd} with momentum 0.9. To induce non-\ac{iid} data distributions across clients, we employ a Dirichlet partitioning scheme with different values of $\alpha$. The divergence hyperparameter $\beta$ (used in the $\lambda$ of the geometrical divergence) is fixed to 1 in all experiments to avoid additional tuning complexity. Each client performs a single local epoch per communication round, in order to keep the experimental setup simple.
For the probe dataset, we select an \ac{iid} subset of 128 samples in each case. All experiments are conducted using Python~v3.10, PyTorch~v2.6, and Flower\footnote{https://flower.ai/}~v1.25, with fixed random seeds to ensure reproducibility.
\subsection{Metric validation}

We first evaluate the validity of the proposed metric. To this end, we focus on the MLP trained on Fashion-MNIST and analyze how the measured divergence evolves as data heterogeneity increases. Specifically, we consider three values of the Dirichlet parameter $\alpha \in \{10, 1, 0.1\}$, ranging from near \ac{iid} to highly heterogeneous regimes. Figure~\ref{fig:exp1} reports the per-client divergence with respect to the global model across communication rounds.

As expected, the first rounds exhibit consistently high divergence across all clients, reflecting the effect of random initialization and early-stage adaptation. As training progresses, we observe a clear stabilization in the near \ac{iid} setting ($\alpha=10$), where all clients converge to similar divergence values. For moderate heterogeneity ($\alpha=1$), the behavior remains stable, although with slightly increased variability across clients.

In contrast, for $\alpha=0.1$, which induces strong heterogeneity, we observe a systematic increase in divergence values, along with the emergence of clients that persistently deviate from the rest. This behavior aligns with our intuition: when local data distributions differ significantly, clients induce distinct geometric organizations of the input space, which is effectively captured by the proposed metric.

\begin{figure*}[!t]
    \centering
    \includegraphics[width=\linewidth]{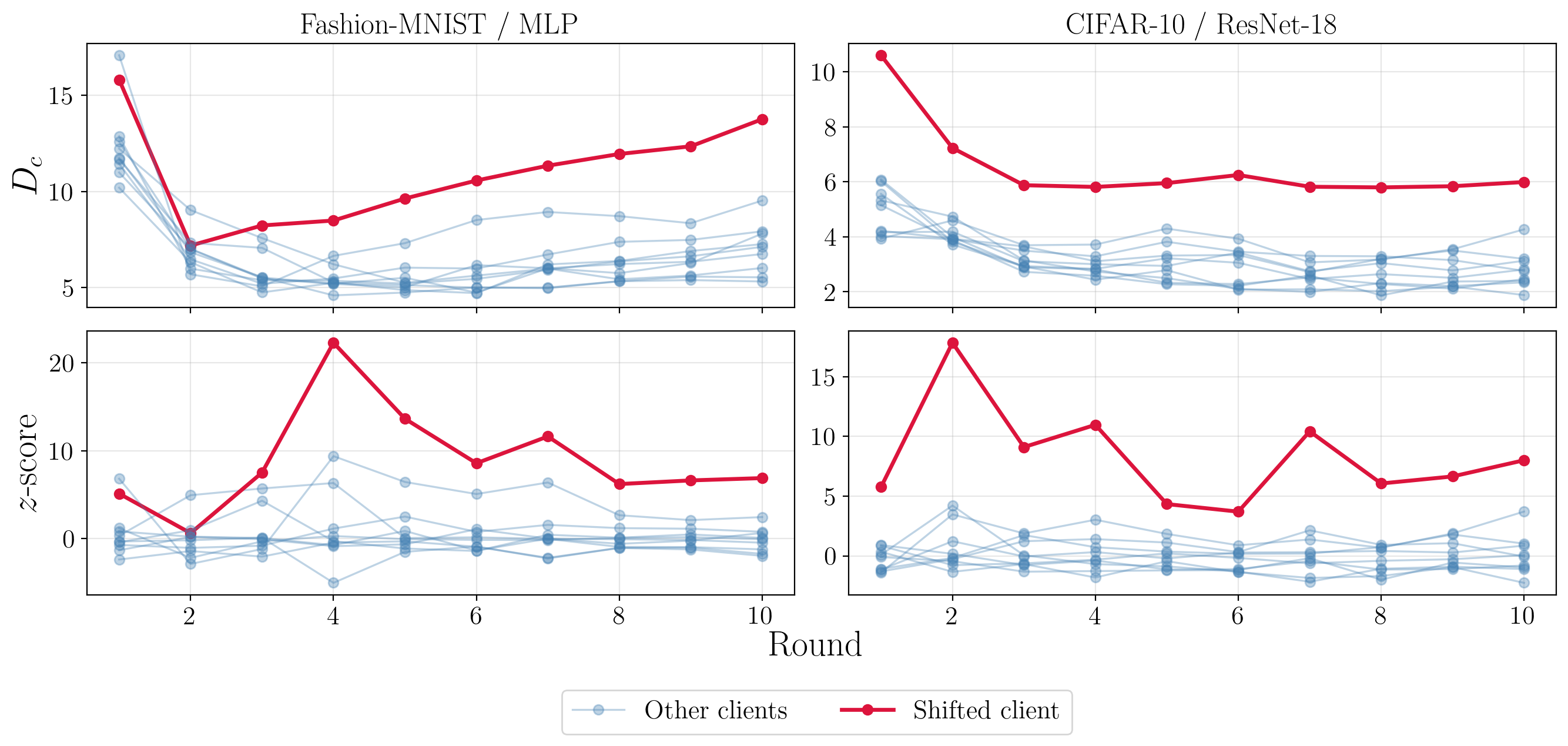}
    \caption{Per client $D_c$ vs $z$-score of the different training rounds. The shifted clients are clearly identifiable, suggesting the validity of the metric to capture these anomalies.}
    \label{fig2:zscore}
\end{figure*}
\subsection{Anomalous client identification}
The second experiment evaluates whether the proposed metric can be used to identify anomalous clients in a standard \ac{fl} setting. In particular, we simulate a \emph{shifted} client, defined as a client whose local data distribution deviates from the rest and may negatively affect the global training process. 

To construct this scenario, we use a Dirichlet partition with $\alpha=1$, ensuring a moderately heterogeneous but realistic setting. Among the $10$ clients, we randomly select one and apply a set of perturbations to its local dataset (e.g., Gaussian noise, rotations, or blurring), thereby simulating a client whose data is corrupted or significantly shifted. The remaining clients follow the original data distribution. Experiments are conducted using both model–dataset pairs described above, with the same training configuration.

To identify such anomalous clients, we define a robust outlier score based on the distribution of client divergences at each communication round. Let $D_c^{(t)}$ denote the geometric divergence of client $c$ with respect to the global model at round $t$. We compute the round-wise median
\[
m^{(t)} = \operatorname{median}_{c}\, D_c^{(t)},
\]
and the corresponding median absolute deviation (MAD)
\[
\operatorname{MAD}^{(t)} = \operatorname{median}_{c}\left( \left| D_c^{(t)} - m^{(t)} \right| \right).
\]

Using these quantities, we define a robust $z$-score for each client:
\[
z_c^{(t)} =
\frac{D_c^{(t)} - m^{(t)}}{\operatorname{MAD}^{(t)} + \varepsilon},
\]
where $0 < \varepsilon \ll1$ ensures numerical stability. Figure~\ref{fig2:zscore} reports both the geometric divergence and the corresponding $z$-scores for all clients, clearly illustrating the effectiveness of the proposed metric. In both scenarios, the shifted client consistently attains significantly higher $z$-scores than the rest, making it easily identifiable as an outlier across communication rounds. In contrast, the remaining clients exhibit low and stable $z$-scores, reflecting consistent behavior within the population.

Moreover, while regular clients tend to stabilize their divergence as training progresses, the shifted client does not converge to the same value or converge in general. Instead, it maintains a persistent deviation from the global model, indicating that its local updates induce a fundamentally different geometric organization of the input space. This behavior is consistent across architectures and datasets, reinforcing the robustness of the proposed approach.

\section{Conclusion}
\label{sec:conclusion}
In this work, we introduced a geometric perspective to characterize client behavior in \ac{fl}. By leveraging activation patterns, we proposed a lightweight metric that measures how local updates modify the functional organization of the model, rather than relying solely on parameter or gradient comparisons.

Through controlled experiments, we showed that the proposed metric effectively captures the impact of data heterogeneity, exhibiting stable convergence in near \ac{iid} settings and increased divergence under stronger non-\ac{iid} regimes. Furthermore, we demonstrated that this geometric signal can be used to reliably identify anomalous clients. In particular, a simple robust $z$-score based on the distribution of client divergences was sufficient to detect shifted clients across different architectures and datasets.

Overall, our results suggest that monitoring the geometric structure induced by \acs{nn} provides a meaningful and interpretable signal to assess client contributions in \ac{fl}. This opens the door to more informed aggregation strategies, where client updates can be weighted or filtered based on their functional behavior.

As future work, we plan to explore tighter integration of this signal into the training loop, as well as its applicability to more complex scenarios, including large-scale federated systems and adversarial settings.

\section*{Acknowledgement}
This research was funded by the projects PID2023-146569NB-C21 and PID2023-146569NB-C22 supported by MICIU/AEI/10.13039/501100011033 and ERDF/UE. Cristian Pérez-Corral received support from the \textit{Conselleria de Educación, Cultura, Universidades y Empleo} (reference CIACIF/2024/412) through the European Social Fund Plus 2021–2027 (FSE+) program of the \textit{Comunitat Valenciana}. Alberto Fernández-Hernández was supported by the predoctoral grant PREP2023-001826 supported by MICIU/AEI/10.13039/501100011033 and ESF+. Jose I. Mestre was supported by the predoctoral grant ACIF/2021/281 of the \emph{Generalitat Valenciana}. Manuel F. Dolz was supported by grant {\small CNS2025-165098} funded by {\small MICIU/AEI/10.13039/501100011033} and by the Plan Gen--T grant {\small CIDEXG/2022/013} of the \emph{Generalitat Valenciana}.

\bibliographystyle{IEEEtran}

\bibliography{bibliography}
\end{document}